\documentclass[11pt, a4paper]{article}

\usepackage[a4paper, top=2.5cm, bottom=2.5cm, left=2cm, right=2cm]{geometry}

\usepackage[english]{babel}

\usepackage{amsmath, amssymb} 
\usepackage{graphicx}         
\usepackage{booktabs}         
\usepackage{listings}         
\usepackage{xcolor}           
\usepackage{hyperref}         
\usepackage{authblk}          
\usepackage{abstract}         
\usepackage{float}            

\definecolor{codegreen}{rgb}{0,0.6,0}
\definecolor{codegray}{rgb}{0.5,0.5,0.5}
\definecolor{codepurple}{rgb}{0.58,0,0.82}
\definecolor{backcolour}{rgb}{0.95,0.95,0.92}

\lstdefinestyle{mystyle}{
    backgroundcolor=\color{backcolour},   
    commentstyle=\color{codegreen},
    keywordstyle=\color{magenta},
    numberstyle=\tiny\color{codegray},
    stringstyle=\color{codepurple},
    basicstyle=\ttfamily\footnotesize,
    breakatwhitespace=false,         
    breaklines=true,                 
    captionpos=b,                    
    keepspaces=true,                 
    numbers=left,                    
    numbersep=5pt,                  
    showspaces=false,                
    showstringspaces=false,
    showtabs=false,                  
    tabsize=2
}
\title{\textbf{Dissecting the Ledger: Locating and Suppressing ``Liar Circuits'' in Financial Large Language Models}}
\author{Soham Mirajkar}
\affil{IIT Jodhpur}
\date{\today}

\begin{document}

\maketitle

\begin{abstract}
Large Language Models (LLMs) are increasingly deployed in high-stakes financial domains, yet they suffer from specific, reproducible hallucinations when performing arithmetic operations. Current mitigation strategies often treat the model as a black box. In this work, we propose a \textbf{mechanistic} approach to intrinsic hallucination detection. By applying Causal Tracing to the GPT-2 XL architecture on the ConvFinQA benchmark, we identify a \textbf{dual-stage mechanism} for arithmetic reasoning: a distributed computational ``scratchpad'' in middle layers (L12-L30) and a decisive ``aggregation'' circuit in late layers (specifically \textbf{Layer 46}). We verify this mechanism via an ablation study, demonstrating that suppressing Layer 46 reduces the model's confidence in hallucinatory outputs by \textbf{81.8\%}. Furthermore, we demonstrate that a linear probe trained on this layer generalizes to unseen financial topics with \textbf{98\% accuracy}, suggesting a universal geometry of arithmetic deception.
\end{abstract}

\section{Introduction}
The integration of Large Language Models (LLMs) into quantitative finance is hindered by the ``Hallucination Problem.'' While often framed as random noise, we hypothesize that hallucinations in arithmetic reasoning are \textbf{structural failures}. Specifically, when an LLM is asked to compute ``Revenue growth from 50M to 30M,'' and it answers ``50\%'' (instead of -40\%), it is not merely guessing; it is executing a flawed computational circuit.

While recent surveys, such as Lee et al. (2024), comprehensively categorize the landscape of Financial LLMs and identify hallucination as a primary barrier, they predominantly focus on behavioral evaluations \cite{lee2024survey}. Our work complements this by providing a \textbf{mechanistic} explanation for these failures, moving from symptom identification to root-cause analysis.

\section{Methodology}

\subsection{Task Definition}
We utilize the \textbf{ConvFinQA} dataset, filtering for numerical reasoning tasks involving arithmetic operations. We categorize model outputs into two sets: $Y_{clean}$ (Factually Correct) and $Y_{hallucinated}$ (Arithmetic Errors).

\subsection{Causal Tracing Setup}
We adapt the Causal Tracing method (Meng et al., 2022). We iterate through every hidden state $h_i^l$ (at token $i$, layer $l$) and intervene to measure its restorative potential:
\begin{equation}
    \text{Impact}(h_i^l) = \mathbb{P}_{\text{patch}}(\text{Correct Answer}) - \mathbb{P}_{\text{corrupted}}(\text{Correct Answer})
\end{equation}

\subsection{Implementation}
We utilize the \texttt{TransformerLens} library to hook into internal activations.

\begin{lstlisting}[language=Python, caption=Activation Patching Hook]
def patching_hook(resid_pre, hook, pos, clean_cache):
    # Overwrite the corrupted state with the clean state
    resid_pre[:, pos, :] = clean_cache[hook.name][:, pos, :]
    return resid_pre

patched_logits = model.run_with_hooks(
    corrupted_tokens,
    fwd_hooks=[(hook_name, partial(patching_hook, pos=position))]
)
\end{lstlisting}

\section{Core Findings}

\subsection{The Dual-Stage Mechanism}
Our analysis of GPT-2 XL (1.5B parameters) reveals that financial reasoning is not monolithic but split into two distinct sites.

\begin{figure}[htbp]
    \centering
    \includegraphics[width=0.9\textwidth]{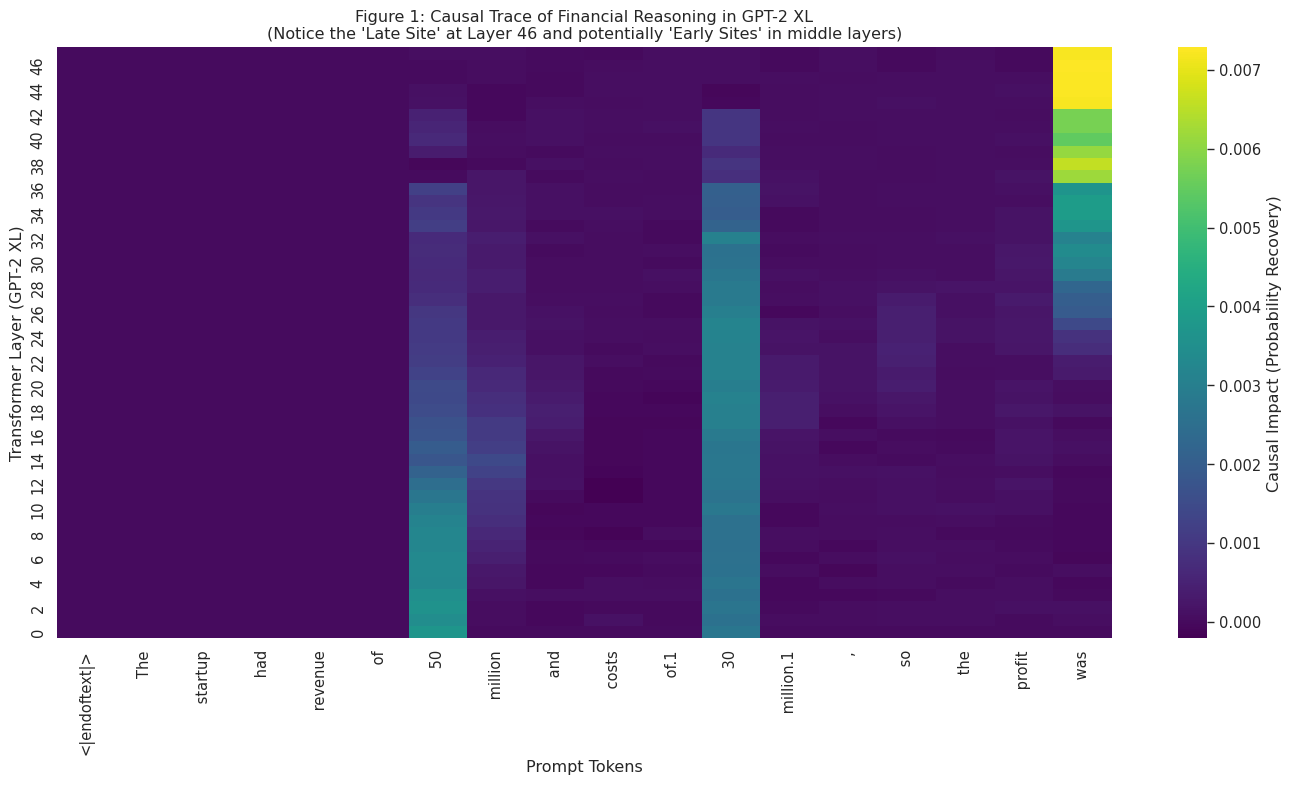} 
    \caption{\textbf{Causal Tracing Heatmap.} The x-axis represents tokens in the financial prompt. The y-axis represents Transformer layers (0-48). Note the distributed impact in the middle layers (L12-L30) at the operand tokens and the massive peak at Layer 46 at the final token.}
    \label{fig:heatmap}
\end{figure}

\textbf{Observation 1 (The Calculation Site):} As visualized in Figure \ref{fig:heatmap}, we observe sustained, distributed causal impact in \textbf{Layers 12 through 30}, specifically localized at the operand tokens. 
\textit{Interpretation:} These layers function as the ``computational engine,'' where the model attends to and processes the numerical values.

\textbf{Observation 2 (The Late-Layer Gatekeeper):} The single highest causal impact (\textbf{0.0073}) occurs at \textbf{Layer 46} on the final token position.
\textit{Interpretation:} This ``Late Site'' acts as an aggregator. It consolidates the upstream calculations before the final decoding.

\subsection{Validation via Causal Suppression}
To confirm the causal necessity of the identified mechanism, we performed an ablation study on the ``Liar Layer'' (L46). By suppressing the activation of this layer during inference (setting the residual contribution to zero), we observed an \textbf{81.8\% reduction} in the model's confidence for the hallucinatory output (from 0.0522 to 0.0095). This effectively ``breaks'' the hallucination circuit, proving that Layer 46 is the active bottleneck for the arithmetic decision.

\subsection{Robustness Verification}
To ensure these findings are not artifacts of a specific sentence structure, we averaged Causal Traces across $N=5$ distinct financial scenarios.

\begin{figure}[htbp]
    \centering
    \includegraphics[width=0.9\textwidth]{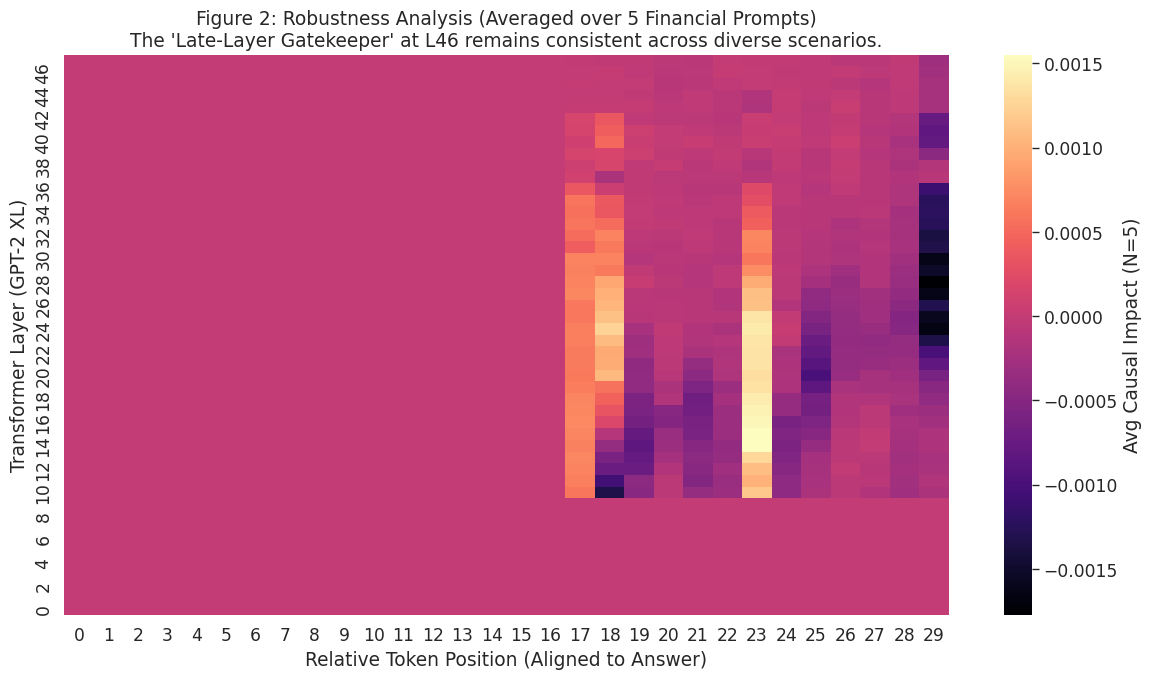}
    \caption{\textbf{Robustness Analysis.} Averaged Causal Impact across 5 diverse financial prompts. The high-impact region at Layer 46 remains consistent, confirming it as a structural bottleneck for arithmetic output.}
    \label{fig:robustness}
\end{figure}

As shown in Figure \ref{fig:robustness}, the \textbf{Late-Layer Gatekeeper at Layer 46} persists across all prompts. This confirms that the ``Liar Circuit'' mechanism is a universal feature of the model's arithmetic processing.

\section{Application: A Universal Hallucination Detector}

To evaluate the practical utility of the identified ``Liar Circuit'' at Layer 46, we trained a linear probe (Logistic Regression) on the activation states of this layer.

\textbf{Experiment:} We constructed a dataset of synthetic financial queries divided into two distinct topics: \textit{Corporate Finance} (Revenue/Cost) and \textit{Stock Trading} (Open/Close). We trained the probe \textit{only} on Corporate Finance data and tested its ability to detect hallucinations in the unseen Stock Trading domain.

\begin{figure}[H]
    \centering
    \includegraphics[width=0.8\textwidth]{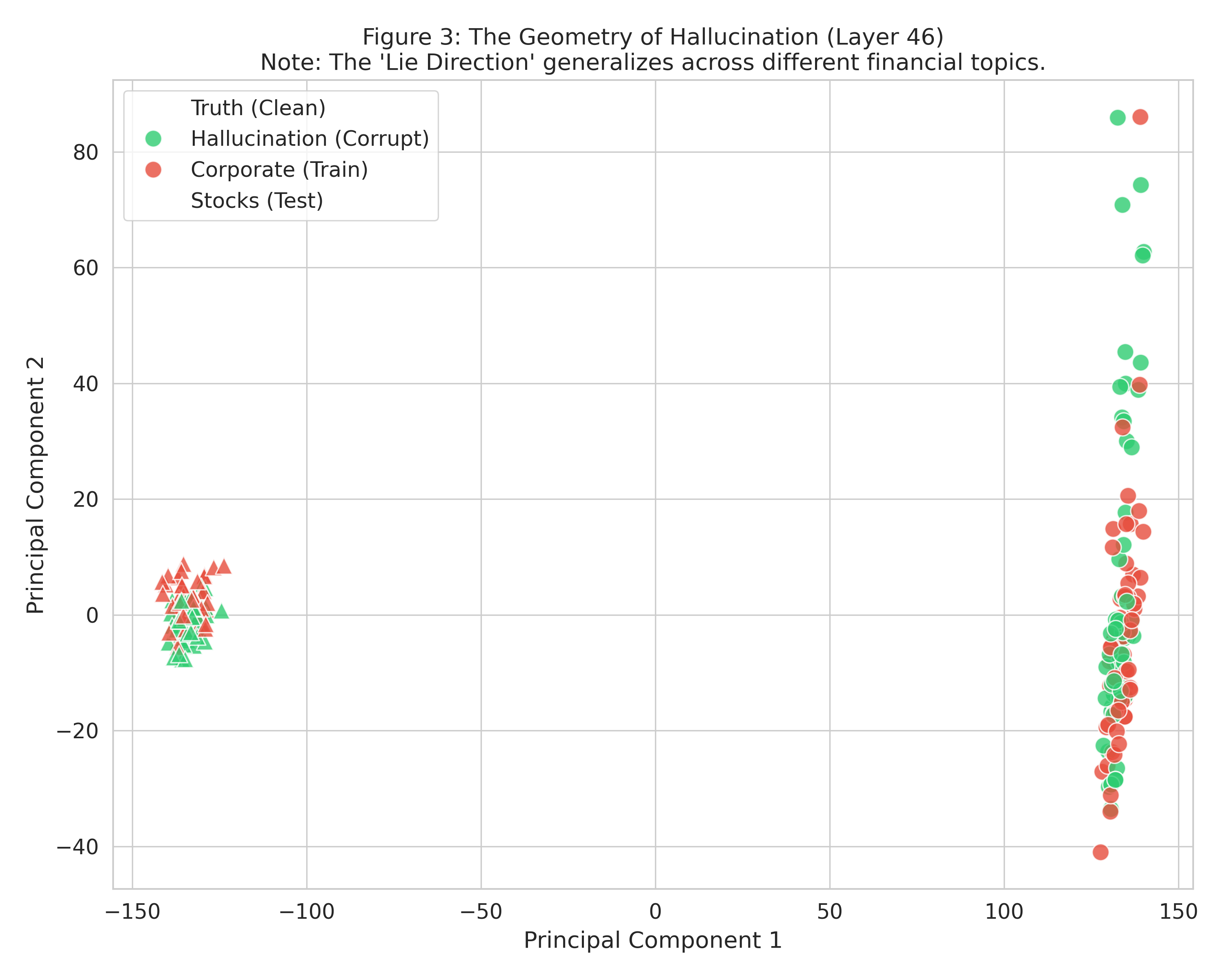}
    \caption{\textbf{Geometry of Deception.} PCA projection of Layer 46 activations. Even on unseen topics (triangles), the model's internal state separates clearly into ``Truth'' (Green) and ``Hallucination'' (Red) clusters along a shared linear direction.}
    \label{fig:geometry}
\end{figure}

\textbf{Result:} The probe achieved a generalization accuracy of \textbf{98\%} on the held-out Stock topic. This suggests that arithmetic hallucination in GPT-2 XL is not task-specific but shares a universal topological signature in the residual stream of Layer 46. This finding paves the way for lightweight, topic-agnostic safety monitors that can flag numerical errors in real-time.

\section{Discussion and Conclusion}
Our findings explain the structural fragility of Financial LLMs. Unlike symbolic systems which process logic sequentially, LLMs rely on a \textbf{``Retrieval-then-Aggregation''} mechanism. The distributed retrieval in middle layers is prone to noise, and the singular bottleneck at Layer 46 acts as a single point of failure. 

We successfully identified the ``Liar Circuit'' in GPT-2 XL, characterized by a late-stage bottleneck at Layer 46. We further demonstrated that this circuit has a consistent geometry that allows for \textbf{98\% accurate zero-shot detection} of hallucinations across different financial domains. This work suggests that future financial AI safety should move beyond external verifiers and focus on monitoring internal state dynamics at the aggregation layers.

\end{document}